\documentclass[letterpaper, 10 pt, conference]{ieeeconf}  % Comment this line out if you need a4paper

\IEEEoverridecommandlockouts                              % This command is only needed if 
\usepackage{graphics} % for pdf, bitmapped graphics files
\usepackage{epsfig} % for postscript graphics files
\usepackage{times} % assumes new font selection scheme installed
\usepackage{amsmath} % assumes amsmath package installed
\usepackage{amssymb}  % assumes amsmath package installed

\usepackage{amsmath}
\usepackage{comment}
\usepackage{url}
\usepackage{color}
\usepackage{amsfonts}
\usepackage{mathtools}
\usepackage{cite}
\usepackage{bm}
\usepackage{algorithm}
\usepackage{algorithmic}
\usepackage{multirow}
\usepackage{balance}

\usepackage{color}
\usepackage{booktabs}

\def\ie{\textit{i.e}\onedot}

\def\etal{\textit{et al}\onedot}

\makeatletter
\let\MYcaption\@makecaption
\makeatother

\usepackage[font=footnotesize]{subcaption}

\makeatletter
\let\@makecaption\MYcaption
\makeatother

\usepackage{subcaption}

\def\ie{{\it i.e.}}
\def\etal{{\it et al. }}

\title{\LARGE \bf
Spline-Interpolated Model Predictive Path Integral Control\\with Stein Variational Inference for Reactive Navigation
}
\author{Takato Miura$^{1}$, Naoki Akai$^{1,2}$, Kohei Honda$^{1}$, and Susumu Hara$^{1}$% <-this % stops a space
\thanks{*This work was supported by KAKENHI under Grant 23K03773.}% <-this % stops a space
\thanks{$^{1}$Takato Miura, Naoki Akai, Kohei Honda, and Susumu Hara are with the Graduate School of Engineering, Nagoya University, Nagoya 464-8603, Japan}%
\thanks{$^{2}$Naoki Akai is with the LOCT Co., Ltd., Nagoya 464-0805, Japan}%
\thanks{E-mail of corresponding author: {\tt\small akai@nagoya-u.jp}}
}

\newcommand{\argmax}{\mathop{\rm argmax}\limits}
\newcommand{\argmin}{\mathop{\rm argmin}\limits}

\begin{document}

\newcommand{\1}{\mbox{1}\hspace{-0.25em}\mbox{l}}
\renewcommand{\baselinestretch}{1.0}

\maketitle
\thispagestyle{empty}
\pagestyle{empty}

%%%%%%%%%%%%%%%%%%%%%%%%%%%%%%%%%%%%%%%%%%%%%%%%%%%%%%%%%%%%%%%%%%%%%%%%%%%%%%%%

\begin{abstract}
This paper presents a reactive navigation method that leverages a Model Predictive Path Integral (MPPI) control enhanced with spline interpolation for the control input sequence and Stein Variational Gradient Descent (SVGD). The MPPI framework addresses a nonlinear optimization problem by determining an optimal sequence of control inputs through a sampling-based approach. The efficacy of MPPI is significantly influenced by the sampling noise. To rapidly identify routes that circumvent large and/or newly detected obstacles, it is essential to employ high levels of sampling noise. However, such high noise levels result in jerky control input sequences, leading to non-smooth trajectories. To mitigate this issue, we propose the integration of spline interpolation within the MPPI process, enabling the generation of smooth control input sequences despite the utilization of substantial sampling noises. Nonetheless, the standard MPPI algorithm struggles in scenarios featuring multiple optimal or near-optimal solutions, such as environments with several viable obstacle avoidance paths, due to its assumption that the distribution over an optimal control input sequence can be closely approximated by a Gaussian distribution. To address this limitation, we extend our method by incorporating SVGD into the MPPI framework with spline interpolation. SVGD, rooted in the optimal transportation algorithm, possesses the unique ability to cluster samples around an optimal solution. Consequently, our approach facilitates robust reactive navigation by swiftly identifying obstacle avoidance paths while maintaining the smoothness of the control input sequences. The efficacy of our proposed method is validated on simulations with a quadrotor, demonstrating superior performance over existing baseline techniques.
\end{abstract}

\section{Introduction\label{Intro}}

\begin{comment}
\begin{figure}[t]
    \centering
    \includegraphics[width=1.0\linewidth]{figure/trajectory_graph_Ours_revise.pdf}
    \caption{ 
A selected simulation result. The original MPPI's candidate paths (light green) are erratic due to the randomness of the injected Gaussian noise, leading to a jerky predicted path (green). To achieve a smoother solution, we employ spline interpolation in MPPI, which samples sparse control points and interpolates them using spline curves. This approach enables the generation of smoother candidate paths. However, some paths (light purple) collide with obstacles due to the scarcity of control points. To address this, we apply the SVGD method, enhancing MPPI with spline interpolation and SVGD, referred to as SCP-MPPI. This method improves the quality of collision-free interpolated samples (light blue). Consequently, SCP-MPPI is able to identify a smooth and collision-free solution (blue), even with sparse control points, resulting in improved computational efficiency. }
    \label{fig:mppi_task}
    % \vspace{-6mm}
\end{figure}
\end{comment}

\begin{figure}[!t]
    \begin{center}
        \includegraphics[width = 90 mm]{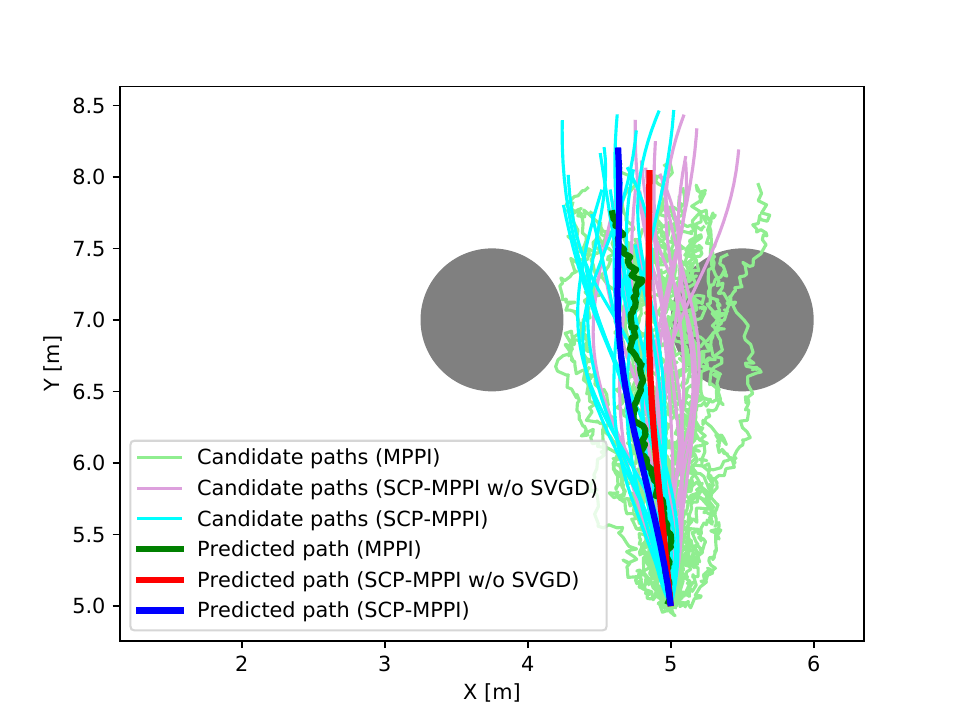}
        \caption{ A selected simulation result. The original MPPI's candidate paths (light green) are erratic due to the randomness of the injected Gaussian noise, leading to a jerky predicted path (green). To achieve a smoother solution, we employ spline interpolation in MPPI, which samples sparse control points and interpolates them using spline curves. This approach enables the generation of smoother candidate paths. However, some paths (light purple) collide with obstacles due to the scarcity of control points. To address this, we apply the SVGD method, enhancing MPPI with spline interpolation and SVGD, referred to as SCP-MPPI. This method improves the quality of collision-free interpolated samples (light blue). Consequently, SCP-MPPI is able to identify a smooth and collision-free solution (blue), even with sparse control points, resulting in improved computational efficiency. }
        \label{fig:mppi_task}
    \end{center}
\end{figure}

\begin{comment}
Unmanned Aerial Vehicles (UAVs) have attracted significant attention due to their potential to contribute to a variety of applications, especially in unknown and challenging environments.
In such environments with many unforeseen obstacles, UAVs require the capability of reactive point-to-goal navigation, where UAVs navigate to a given destination while avoiding unforeseen obstacles.
To achieve agile and robust reactive navigation, local motion planners need to find smooth and collision-free trajectories with less number of random samples, which leads to less computational load.
\end{comment}

Unmanned Aerial Vehicles (UAVs) have garnered considerable interest for their capability to enhance a wide range of applications, particularly in unfamiliar and demanding settings.
In these environments, characterized by numerous unexpected obstacles, UAVs must have the ability for reactive point-to-goal navigation, enabling them to reach designated destinations while bypassing unforeseen obstacles.
To facilitate agile and reliable reactive navigation, local motion planners are tasked with identifying smooth and collision-free paths using a minimal number of random samples, thereby reducing computational demands.

\begin{comment}
Among various practical local motion planning methods~\cite{DWA, SLP, CEM}, Model Predictive Path Integral control (MPPI)~\cite{MPPI2016,MPPI2017,MPPI2018} is relatively computationally efficient and is a sampling-based stochastic Model Predictive Control (MPC)~\cite{SMPC,SMPC2} framework that can handle non-linearity and non-differentiability of the environment, such as those involving UAV's dynamics and cost maps.
MPPI can find a smooth control input sequence by sampling a large enough number of samples and taking a weighted average of them.
In the MPPI algorithm, the variance of the random samples is an important factor to improve collision avoidance performance.
On the other hand, for the solution smoothness, we need to spread the ``large enough'' number of samples, but the greater the sample variance, the greater the number of samples required.
Since the number of samples has a limitation due to hardware settings, we need to compromise on the sample variances.
Therefore, how to find a smooth solution with an insufficient number of samples is a challenging issue.
\end{comment}

Among various practical local motion planning methods, such as Dynamic Window Approach (DWA), Sampling-based Local Planner (SLP), and Cross-Entropy Method (CEM)~\cite{DWA, SLP, CEM}, Model Predictive Path Integral (MPPI) control~\cite{MPPI2016,MPPI2017,MPPI2018} stands out for its relative computational efficiency.
MPPI, a sampling-based stochastic Model Predictive Control (MPC) framework~\cite{SMPC,SMPC2}, is adept at handling the complexities of non-linear and non-differentiable environments, including UAV dynamics and cost maps.
This method is capable of finding a smooth control input sequence by sampling a sufficiently large number of samples and taking their weighted average.
In the MPPI algorithm, the variance of the random samples is crucial for improving collision avoidance, whereas solution smoothness necessitates spreading a ``large enough'' number of samples across a broad variance.
However, a larger sample variance increases the number of samples required, which is constrained by hardware limitations.
Therefore, finding a smooth solution with an insufficient number of samples due to these limitations poses a significant challenge.

To find a smooth solution even with the insufficient number of samples, one key idea is to utilize sample interpolation between the time dimensions.
That is, we can obtain a smooth solution by interpolating a sparse predicted control input sequence (hereafter referred as to \emph{sparse control points}).
However, the sparse control points decrease the predictive resolution of the dynamics, resulting in decreasing the number of feasible (\ie, collision-less) samples, resulting in degrading of the smoothness and sample efficiency.

\begin{comment}
In this paper, we propose an advanced MPPI algorithm, named Sparse Control Points MPPI (SCP-MPPI), to enhance the smoothness and collision-less of the MPPI solution with the sparse control points.
In the proposed method, we first incorporate the spline interpolation of the control points into the standard MPPI algorithm to find a smooth trajectory even if the control points are sparse.
With the spline interpolation, we can get smooth sample trajectories with sparse control points, but most of the samples become infeasible because of the fewer control points.
To address the collisions in the interpolated parts, we then utilize the Stein Variational Gradient descent (SVGD) method~\cite{SVGD2016} after the interpolation.
The SVGD method can increase the feasible samples by transporting them based on the gradient of the optimal distribution.
As a result, SCP-MPPI can find a smooth and collision-less trajectory compared to the standard MPPI method, as shown in Fig.~\ref{fig:mppi_task}.
\end{comment}

In this paper, we propose an enhanced Model Predictive Path Integral (MPPI) algorithm, named Sparse Control Points MPPI (SCP-MPPI), designed to improve the smoothness and collision-free performance of MPPI solutions.
Our approach first integrates spline interpolation of control points into the conventional MPPI framework to generate smooth control input sequences, even when control points are sparse.
While spline interpolation allows for smoother sample trajectories, it often results in a majority of samples being infeasible due to the reduced number of control points, for example, some predicted paths collide with obstacles.
To mitigate collisions within the interpolated sequences, we subsequently apply the Stein Variational Gradient Descent (SVGD) method~\cite{SVGD2016} following interpolation.
SVGD enhances the number of feasible samples by transporting them in accordance with the gradient of the optimal distribution.
Consequently, SCP-MPPI is capable of producing smooth and collision-free trajectories, demonstrating superior performance compared to the standard MPPI approach, as illustrated in Fig.~\ref{fig:mppi_task}.

\begin{comment}
To summarize the main contribution of this work, we first introduce a novel MPPI-based method, named SCP-MPPI.
This method integrates spline interpolation with the SVGD algorithm, enabling smooth and collision-free motion even with sparse control points.
Consequently, the computational complexity associated with random sampling and evaluation for finding the optimal solution is significantly reduced due to the sparsity of the control points.
In our experiments, we demonstrate the superior efficiency of our proposed method by achieving comparable performance with significantly fewer control points.
Whereas conventional methods required up to 150 points, our approach yields sufficient results using only 4 points.
%Our extensive
In addition, we demonstrate that the SCP-MPPI outperforms other baseline methods in terms of success rates of obstacle avoidance and average speed of the UAV, resulting in reduced flight times to reach the destination.
These results suggest that our method can achieve robust and agile control for reactive navigation of UAVs.
\end{comment}

To encapsulate the primary contributions of this study, we introduce a novel method based on the MPPI framework, named SCP-MPPI.
This approach combines spline interpolation with the SVGD algorithm to facilitate smooth and collision-free navigation, even though sampled control inputs are sparse.
The utilization of sparse control points significantly diminishes the computational complexity typically associated with the random sampling and evaluation processes necessary for identifying the optimal solution.
Through experimental validation, we reveal the enhanced efficiency of our proposed method, achieving comparable performance levels with fewer control points.
While traditional methods necessitate up to 150 points, our method demonstrates effectiveness with as few as 4 points.
Moreover, SCP-MPPI surpasses other baseline methods in obstacle avoidance success rates and the average speed of the UAV, culminating in shorter flight durations to the designated target.
These findings show the potential of our method to provide robust and agile control for the reactive navigation of UAVs.

\section{Related work}
\label{sec:related_work}

Although MPPI is a promising framework for sampling-based MPC because of its relatively high sampling efficiency, it often struggles with the oscillation of the solution when the number of samples is small due to the limitation of computational resources.
To overcome this limitation, how to improve the sample efficiency is an open issue.
Existing approaches can be broadly categorized into the following two types:

\subsection{Random Sampling in Different Spaces}

One approach to finding a smooth control input sequence with fewer samples is to sample other spaces instead of control inputs.
Kim \etal find smooth control input sequences for autonomous vehicles by sampling in lifted input space, \ie, differential values of control inputs~\cite{kim2022smooth}.
Although this approach is easy to apply, it penalizes the agile behavior of the vehicle, which may degrade obstacle avoidance performance.
Another approach is to sample the random parameters in a geometric space.
Some studies can find a smooth geometric path by sampling parameters of the geometric curves~\cite{higgins2023model, SVGDMPC}, such as spline curves, or using a kernel function~\cite{liu2023collision}.
However, these methods cannot find actual control inputs and require a lower-level controller to track the optimized path.

\subsection{Improving Quality of the Random Samples}

The second approach to finding a more sample-efficient solution is to increase the number of low-cost and feasible random samples.
Some studies propose ancillary controllers that induce the MPPI solution to far from infeasible area~\cite{balci2022constrained, williams2018robust}.
These methods often struggle with the nonlinearity of the system to formulate the ancillary controllers.
Other studies try to improve the prior distribution to reduce the infeasible samples~\cite{yin2022trajectory, sacks2023learning, asmar2023model, mohamed2023towards}.

As a more advanced method, variational inference techniques have been attracting attention to handle more representative distribution, as their prior~\cite{okada2020variational, wang2021variational, barcelos2021dual, SVGDMPC}.
In particular, Stein Variational MPC (SV-MPC)~\cite{SVGDMPC} can transport the random samples directly to make them feasible and low-cost by using SVGD~\cite{SVGD2016}. 
However, the computational complexity per samples is expensive, and requires reducing the control points.

Our proposal in this work restricts the sampling of control input sequences to lower dimensions and approximates the intermediate points using spline interpolation.
This approach allows multi-dimensional control input sequence optimization problems to be approximated with lower-dimensional control input sequences, thus allowing more efficient trajectory sampling.
By updating the sampled trajectories using the SVGD method, the proposed method achieves effective trajectory updates in response to the environment.
The proposed approach is expected to overcome the problem of falling into local minima by exploiting SVGD's ability to promote convergence among solutions and to encourage the solutions to spread evenly over the entire posterior distribution, aided by the gradient information of the cost function.

\section{ALGORITHM REVIEW}

\subsection{MPPI Algorithm Review}
\label{sec:MPPI_rev}

In this section, we briefly describe the basic algorithm of MPPI.
The MPPI theory considers a general discrete-time dynamical system for the controlled object.
The control input sequence $U = ({\bf u}_{0},~{\bf u}_{1},~\ldots,~{\bf u}_{T-1})$ represents the control inputs from the current time point to future $T$ time points and controls the system, where ${\bf u}$ is a control point.
The MPPI algorithm models the system as, ${\bf x}_{t+1} = {\bf F}( {\bf x}_{t}, {\bf v}_t)$, 
where $ {\bf x}_{t} \in \mathbb{R}^n$ represents the state of the system at time $t$, and ${\bf v}_{t}\in \mathbb{R}^m$ is the input to the system at time $t$.
We assume that ${\bf F}$ can be nonlinear and non-differentiable.
In practice, the control input sequence is subject to noise, and the actual control input sequence is represented as $V = ({\bf v}_{0},~{\bf v}_{1},~\ldots,{\bf v}_{T-1})$, where ${\bf v}_{t} \sim \mathcal{N}({\bf u}_{t}, \Sigma)$, and $\Sigma$ is the covariance matrix of the Gaussian distribution.

MPPI aims to approximate the action distribution $q$ using randomly generated $K$ control input sequences $V^{(k)}$. The probability density function of the action distribution $q$ is defined as, 
\begin{align}
q(V) = Z^{-1} \prod_{t=0}^{T-1}\exp\left(-\frac{1}{2}({\bf v}_t-{\bf u}_t)^\top\Sigma^{-1}({\bf v}_t-{\bf u}_t) \right),\label{q_distribution}
\end{align}
where $Z = \left(\sqrt{(2\pi)^m|\Sigma|}\right)$ is the normalization term.
To find the optimal $U^*$, MPPI solves the following stochastic optimization problem,
\begin{align}
    U^{*} = \argmin_{U} \mathbb{E}_{q} \left[ \psi({\bf x}_{T}) + \sum_{t=0}^{T-1} \mathcal{L}({\bf x}_t,~{\bf  u}_t) \right],  \label{original_opt_problem}
\end{align}
where $\psi(\cdot)$ is the terminal cost function, and $\mathcal{L}(\cdot)$ is the stage cost function. 
In the following, we desire the sequence cost of the $k$th trajectory as,
\begin{align}
    \tilde{S}_k =  \psi({\bf x}^{(k)}_T) + \sum_{t=0}^{T-1} \mathcal{L}({\bf x}^{(k)}_t,~{\bf  u}_t^{(k)}).\label{cost_MPPI} 
\end{align}
Since directly solving the stochastic optimization problem shown in equation~(\ref{original_opt_problem}) is challenging, MPPI instead aims to minimize the Kullback-Liebler (KL) divergence between $q$ and the optimal optimality likelihood $\mathbb{Q^*}$:
\begin{align}
    U^{*} = \argmin_{U} \text{KL}(q^{*} \| q).
    \label{MPPI_KL}
\end{align}
To minimize the KL divergence, the right-hand term of equation (\ref{MPPI_KL}) can be re-written as:
\begin{align}
U^* &= \argmax_{U} \mathbb{E}_{q^*} \left[ \frac{1}{2}\sum_{t=0}^{T-1}({\bf v}_{t} - {\bf u}_{t})^\top \Sigma^{-1} ({\bf v}_{t} - {\bf u}_{t})\right], \nonumber \\
& = \mathbb{E}_{q^*} [V^{(k)}]. \label{opt_u}
\end{align}
The optimal control input sequence $U^*$ is obtained by sampling $V^{(k)}$ from the optimal optimality likelihood $\mathbb{Q^*}$ and taking its expectation value in equation~(\ref{opt_u}). 
However, since direct sampling from $\mathbb{Q^*}$ is not possible, MPPI approximates the optimal optimality likelihood using the importance sampling technique~\cite{kloek1978bayesian} as,
\begin{align}
& U^* = \mathbb{E}_{q} \left[\frac{q^*(V)}{q(V^{(k)})} V^{(k)}\right] \simeq \sum_{k=0}^{K-1} w\left(V^{(k)}\right) V^{(k)}, \label{opt_u_by_is}\\
& w\left(V^{(k)}\right) = \frac{1}{\eta} \exp \left( - \frac{1}{\lambda}\tilde{S} (V^{(k)} + \lambda \sum_ {t=0}^{T=1}({\bf \hat{u}}_t - {\bf \tilde{u}}_t)^\top\Sigma^{-1}{\bf v}_t)\right)\label{weight},
\end{align}
where $w(\cdot)$ is the weight function and ${\bf \hat{u}}_t$ represents the previous optimal solution, and ${\bf \tilde{u}}_t$ represents a nominal control input sequence applied to the system.
The MPPI algorithm leverages the law of large numbers to minimize the KL divergence by averaging a sufficient number of weighted samples, obtaining the optimal solution without iterative solution updates.

\subsection{SVGD Algorithm Review}

To address the task of minimizing the KL divergence as shown in equation~(\ref{MPPI_KL}) and to identify an appropriate distribution $q$, we utilize the SVGD method. Similar to MPPI, SVGD aims to approximate the optimal distribution $q^*$. This method employs particles to represent the optimal action distribution $q^*$. In this work, the particles symbolize vectors of the difference of control input sequences, which are defined as $\Delta U = (\Delta {\bf u}_0,~\Delta {\bf u}_1,~\ldots,\Delta {\bf u}_{T-1})$. During the SVGD process, these particles are adjusted along the gradient direction of optimality likelihood $p$ to converge with the optimal distribution $q^*$. The optimality likelihood $p$ is a modeled version of the optimal distribution, formulated based on the optimal control problem. Specifically, the particle update process is performed iteratively based on following equations:
\begin{align}
    &\Delta U^{(k)} \leftarrow \Delta U^{(k)} + \epsilon\Phi{(\Delta U^{(k)}}),\label{SVGD_transport}\\
    &\Phi = \argmax_{\Phi \in \mathcal{F}} -\nabla_{\epsilon} \text{KL}(q \| p \simeq q^*),\label{Stein Discrepancy}
\end{align}
where the step size $\epsilon$ is provided as a hyperparameter, and $\mathcal{F}$ is a set of functions with bounded Lipschitz norms.
Solving the variational problem shown in equation~(\ref{Stein Discrepancy}) involves the choice of a smooth function $\Phi$, which affects the computational cost. 
However, to broaden $\mathcal{F}$, requiring an infinite number of basis functions makes solving the variational problem in Stein Discrepancy computationally expensive \cite{Stein_discrepancy}.
Therefore, the function $\Phi$ is considered to reside in the unit ball of a Reproducing Kernel Hilbert Space (RKHS).
This RKHS takes the form of $\mathcal{H} = \mathcal{H}_0 \times \dots \mathcal{H}_0$.
$\mathcal{H}_0$ is a reproducing kernel Hilbert space that holds scalar-valued functions, and the kernel function used to characterize the properties of functions in this space is $k(\Delta U',~\Delta U)$.
This kernel function defines the relationship and inner product between different particles $\Delta U'$ and $\Delta U$ within $\mathcal{H}_0$, capturing the similarity between particles.

In other words, $\Phi$ is expressed as follows:
\begin{align}
    \Phi = \argmax_{\Phi \in \mathcal{H}} -\nabla_{\epsilon} \text{KL}(q \| p).
\end{align}
The function $\Phi$ characterizes the optimal perturbation that maximally reduces the KL divergence. 
Based on this, particles are updated following the approximate steepest descent direction $\Phi{(\Delta U)}$:
\begin{align}
    \Phi{(\Delta U)} &= \frac{1}{K}\sum_{j=0}^{K-1} \left[k(\Delta U^{(j)},~\Delta U)\nabla_{\Delta U^{(j)}}\log(p(\Delta U^{(j)}) \right. \nonumber \\
    &\quad + \left. \nabla_{\Delta U^{(j)}}k(\Delta U^{(j)},~\Delta U)\right]. \label{SVGD_}
\end{align}
The first term in equation~(\ref{SVGD_}) represents a force acting on the particle approximation of the optimality likelihood in the direction of the gradient. This term utilizes the gradient of the optimality likelihood to attract particles towards the optimality likelihood.

On the other hand, the second term serves as a repulsive force. It pushes particles away from each other when they get too close. This repulsive term helps prevent particles from converging to local optima, allowing the method to avoid collapsing into a unimodal distribution.

Unlike conventional Monte Carlo methods, this approach uses gradient-based forces to control interactions between particles. Even when using only a single particle, this method aims to maximize $\log(p(\Delta U^{(j)}))$ using gradient information. This is because the repulsive term vanishes, resulting in $\nabla_{\Delta U}k(\Delta U,~ \Delta U) = 0$.

\section{SPARSE CONTROL POINTS MPPI}

\subsection{Overview of the Proposed Method}

\begin{figure*}[!t]
    \centering
    \includegraphics[width=1.0\linewidth]{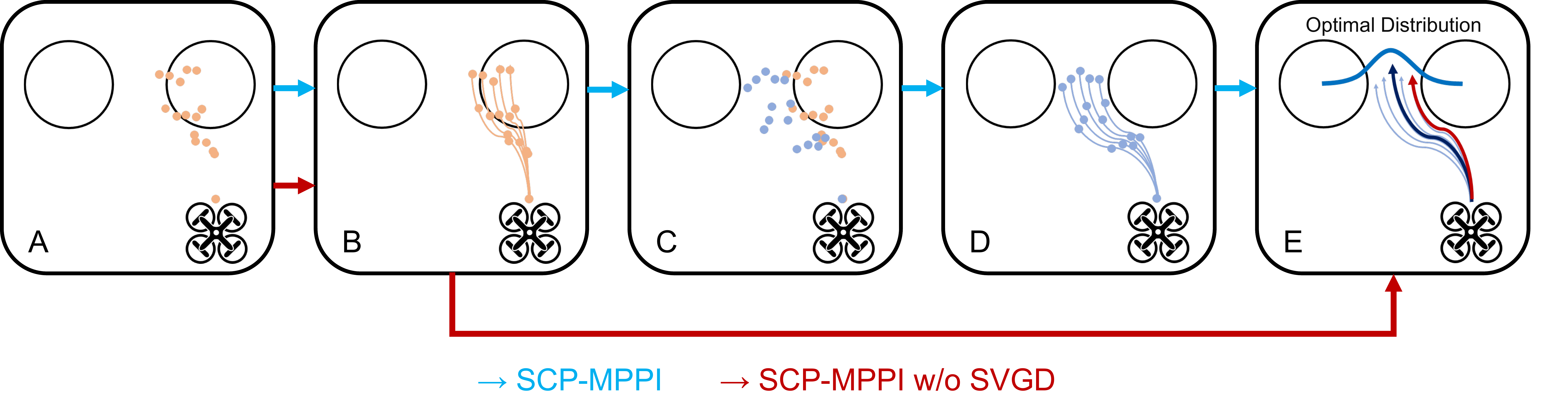}
    \caption{ The overview of SCP-MPPI with and without the SVGD method. (A) Initially, SCP-MPPI sparsely samples control points (illustrated as light red points) and (B) employs spline curves to interpolate these points, creating control input sequences (shown as light red lines). (C) If SVGD is applied, the sparse control points are adjusted (depicted as light blue points), followed by (D) another round of spline interpolation (resulting in light blue lines). (E) The process culminates in determining the optimal action distribution and generating a reactive navigation path (represented by a blue line). While SCP-MPPI without SVGD can navigate around obstacles (indicated by a red line), incorporating SVGD enables SCP-MPPI to approximate the global optimum more closely, even when the initial distribution is not close to the optimal solution. }
    \label{fig:flow_SPC-MPPI}
\end{figure*}

\begin{comment}
This subsection details overview of the proposed method.
Figure~\ref{fig:flow_SPC-MPPI} illustrates the overview of the proposed method, named Sparse Control Points MPPI (SCP-MPPI) with and without the SVGD method.
SCP-MPPI starts by sampling sparse control points based on the control points from a previous time step, similar to MPPI.
We then interpolate the sparse control points using the spline interpolation method to obtain smooth candidates of the control input sequences.
However, the spline-interpolated samples are not close to samples drawn from the optimal action sequence distribution when the initial distribution is far from the otpimal one because of the sparsity of the control points, which may degrade the quality of the solution of MPPI.
Therefore, we directly transport the spline-interpolated samples using the SVGD method~\cite{SVGD2016} and it enables us to obtain more better solutions.
It should be noted that the computational complexity can be kept to smaller than SV-MPC~\cite{SVGDMPC} because we only transport the sparse control points.
Finally, SCP-MPPI finds a collision-free and smooth solution by weighting the average of the samples in (\ref{opt_u_by_is}) as shown in Fig.~\ref{fig:flow_SPC-MPPI}.
\end{comment}

This subsection provides an overview of the proposed Sparse Control Points MPPI (SCP-MPPI) method.
Figure~\ref{fig:flow_SPC-MPPI} illustrates the overview of SCP-MPPI with and without the SVGD method.
SCP-MPPI begins by sampling sparse control points from previous time steps, akin to the original MPPI process.
These points are then spline-interpolated to generate smooth control input sequence candidates.
However, due to the sparsity of control points, these spline-interpolated samples might not closely approximate the optimal action sequence distribution, potentially reducing solution quality.
To enhance solution accuracy, we apply SVGD~\cite{SVGD2016} to directly adjust the spline-interpolated samples, leading to improved outcomes.
Notably, this process maintains lower computational complexity compared to SV-MPC~\cite{SVGDMPC} since it only modifies sparse control points.
SCP-MPPI achieves a smooth, collision-free solution by computing the weighted average of these samples, as depicted in Fig.~\ref{fig:flow_SPC-MPPI}.

\subsection{MPPI with spline interpolation}
\label{sec:MPPI_SPLINE}

This subsection details how to incorporate spline interpolation of sparse control points into the MPPI algorithm.
We refer this method to as SCP-MPPI w/o SVGD in this paper.
This MPPI allows us to find smooth trajectories even when the control points are sparse.

In MPPI, the optimal control input sequence $U^{*}$ is obtained via equation~(\ref{opt_u_by_is}) and it is used as the reference control input sequence to obtain $k$th control input sequence.
The $k$th control input sequence $U^{(k)}$ is obtained by adding the Gassusian noises to $U^{*}$.
Then, the generated sequences are evaluated and the optimal sequence is obtained.
MPPI iteratively performs these processes.

SCP-MPPI samples the sparse control input points $\tilde{U} \in \mathbb{R}^{m \times M}$ and adds Gaussian noises to them to generate $k$th sparse control points $\tilde{U}^{(k)}$, where $M$ is the number of sparse control input.
Spline interpolation is performed to the sparse control input points and the $k$th interpolated result is denoted as $U^{(k)}$ that is one control input sequence.
The cost evaluation is executed using the interpolated sequence and the optimal sequence is obtained as the weighted average of them.
$\tilde{U}^{*}$, that is the sparse control inputs generating the optimal input sequence, is extracted from the optimal sequence and is fed to the next inference.

\subsection{Transport Interpolated Sparse Control Points by SVGD}
\label{sec:SPC-MPPI}

\begin{comment}
SCP-MPPI w/o SVGD can avoid obstacles; however, it cannot generate a path close to the optimal one when the initial action sequence distribution is far from the optimal distribution.
To generate the path close to the optimal one in such a case, we employ the SVGD method.

The SVGD method iteratively updates the sampled $M$ control inputs via update of the control input noises $\Delta\tilde{U}^{(k)}$ using equations~(\ref{SVGD_transport}) and (\ref{SVGD_}).
Then, spline interpolation is applied to the updated control inputs again, and evaluation using equation~(\ref{cost_MPPI}) is executed.
Finally, the weighted control inputs are calculated again using equation~(\ref{weight}) to obtain the actual control input for the robot.

The advantage of the SVGD method is directly transporting candidates of the solution, called particles, while considering the posterior distribution.
This enables to move the particles close to the optimal actions even when the particles are sampled at far region.
Consequently, SCP-MPPI can correspond to a case where the optimal distribution quickly changes such as when unforeseen obstacles appear.
\end{comment}

SCP-MPPI w/o SVGD can navigate around obstacles but struggles to generate a path close to the optimal one when the initial action sequence distribution deviates from the optimal distribution.
To address this, we incorporate the SVGD method, which iteratively refines the sampled $M$ control inputs by updating the control input noises $\Delta \tilde{U}^{(k)}$ using equations~(\ref{SVGD_transport}) and (\ref{SVGD_}).
Following this update, spline interpolation is reapplied to the refined control inputs, and their cost is assessed with equation~(\ref{weight}).
The process concludes with the recalculation of weighted control inputs to determine the actual control input for the robot.

The SVGD method's strength lies in its ability to directly adjust solution candidates, \ie~particles, by aligning them with the posterior distribution.
This capability ensures that particles, even those initially positioned far from the optimal actions, can be guided closer to them.
Thus, SCP-MPPI becomes particularly effective in dynamic environments where the optimal distribution may rapidly change, such as in the presence of unforeseen obstacles.

\begin{comment}
The pseudo-code of SCP-MPPI is shown in Algorithm~\ref{alg:SPC-MPPI Algorithm}.
The sparse control points are first sampled and spline interpolation is executed.
In the line~\ref{alg:SVGD}, the SVGD method transports the sparse control inputs.
The spline interpolation is executed again to the transported control inputs and the actual control inputs are obtained at the line~\ref{alg:update control inputs sequence}.
\end{comment}

The pseudo-code for SCP-MPPI is provided in Algorithm~\ref{alg:SPC-MPPI Algorithm}.
Initially, sparse control points are sampled by adding the Gaussian noises and subjected to spline interpolation from line~5 to 7.
At line~\ref{alg:SVGD}, the SVGD method is applied to transport the sparse control inputs.
Following this, spline interpolation is performed once more on the transported control inputs.
The process culminates at line~\ref{alg:update control inputs sequence}, where the actual control inputs are finalized, creating an optimized control input sequence.

\begin{figure}[!t]
\begin{algorithm}[H]
    \caption{Sparse Control Points MPPI}
    \label{alg:SPC-MPPI Algorithm}
    \begin{algorithmic}[1]
        \REQUIRE ~ \\
            $K$: Number of samples;\\ 
            $T$: Number of time steps;\\
            $M$: Number of control points;\\
            $L$: Number of iteration for SVGD;\\
            $\tilde{U} =(\tilde{\bf u}_0,~\tilde{\bf u}_1,~\ldots,~\tilde{\bf u}_{M-1})$: Initial control points;\\
            ~ \\
        \WHILE{task not completed}
            \STATE ${\bf x}_{t_0} \gets$GetState();
            \FOR{$k \gets 0$ to $K-1$}
                \STATE ${\bf x}_0^{(k)} \gets {\bf x}_{t_0}$;
                \STATE $\Delta \tilde U ^ {(k)} \gets (\Delta \tilde{\bf u}^{(k)}_0,~\Delta \tilde{\bf u}^{(k)}_1,~\dots, \Delta \tilde{\bf u}^{(k)}_{M-1})$\label{alg:sampling control inputs};
                \STATE $\tilde{U}^{(k)} \gets (\tilde{\bf u}_0 + \Delta {\bf u}^{(k)}_0,~\tilde{\bf u}_1 + \Delta {\bf u}^{(k)}_1,~\dots, \tilde{\bf u}_{M-1} + \Delta{\bf u}^{(k)}_{M-1})$\label{alg:sample control input sequences};
                \STATE $U ^ {(k)} \gets {\rm CubicSpline}(\tilde{U}^{(k)}) $\label{alg:spline interporated};
                %\gets (\delta \bm u^{(k)}_1,\delta \bm u^{(k)}_2,\dots,\delta \bm u^{(k)}_{T-1})
                \FOR{$t \gets 0$ to $T-1$}
                    \STATE ${\bf x}^{(k)}_{t+1} \gets {\bf F}({\bf x}^{(k)}_t,~{\bf u}^{(k)}_{t})$;
                    \STATE $\tilde{S}_k \gets \tilde{S}_k + c({\bf x}^{(k)}_t,~{\bf u}^{(k)}_t)$;
                \ENDFOR
            \ENDFOR
            \FOR{$l \gets 0$ to $L-1$}
                \FOR{$k \gets 0$ to $K-1$}
                    \STATE $\Delta\tilde{U}^{(k)} \gets {\Delta\tilde{U}^{(k)}} + \epsilon\Phi(\Delta\tilde{U}^{(k)})$\label{alg:SVGD}; 
                \ENDFOR
            \ENDFOR
            \FOR{$k \gets 0$ to $K-1$}
                \STATE $U ^ {(k)} \gets {\rm CubicSpline}(\tilde{U}^{(k)}) $\label{alg:spline interporated SVGD};
                %\gets (\delta \bm u^{(k)}_1,\delta \bm u^{(k)}_2,\dots,\delta \bm u^{(k)}_{T-1})
                \FOR{$t \gets 0$ to $T-1$}
                    \STATE ${\bf x}^{(k)}_{t+1} \gets {\bf F}({\bf x}^{(k)}_t,~{\bf u}^{(k)}_{t})$;
                    \STATE $\tilde{S}_k \gets \tilde{S}_k + c({\bf x}^{(k)}_t,~{\bf u}^{(k)}_t)$;
                \ENDFOR
            \ENDFOR
            \FOR{$m \gets 0$ to $M-1$}
              \STATE $\tilde{\bf u}_m^{*} = \tilde{\bf u}_m + \left\lbrack \sum_{k=0}^{K-1} \frac{\exp{(-\frac{1}{\lambda} \tilde{S}_k)} \Delta {\bf u}^{(k)}_m}{\sum_{j=0}^{K-1}\exp{(-\frac{1}{\lambda} \tilde{S}_j})} \right\rbrack$;
            \ENDFOR
            \STATE $U^{*} \gets {\rm CubicSpline}(\tilde{U}^{*}) $\label{alg:update control inputs sequence};
            \STATE SendCommand $({\bf u}_0^{*})$;
            \STATE Feed $\tilde{U}^{*}$ as next initial control points;
            \STATE Update the current state after receiving feedback;
            \STATE Check for task completion;
        \ENDWHILE
    \end{algorithmic}
\end{algorithm}
\end{figure}

\section{Experiments}
\label{sec:Simulation}

\subsection{Navigation Task}

\begin{figure}[t]
    \centering
    \includegraphics[width=1.0\linewidth]{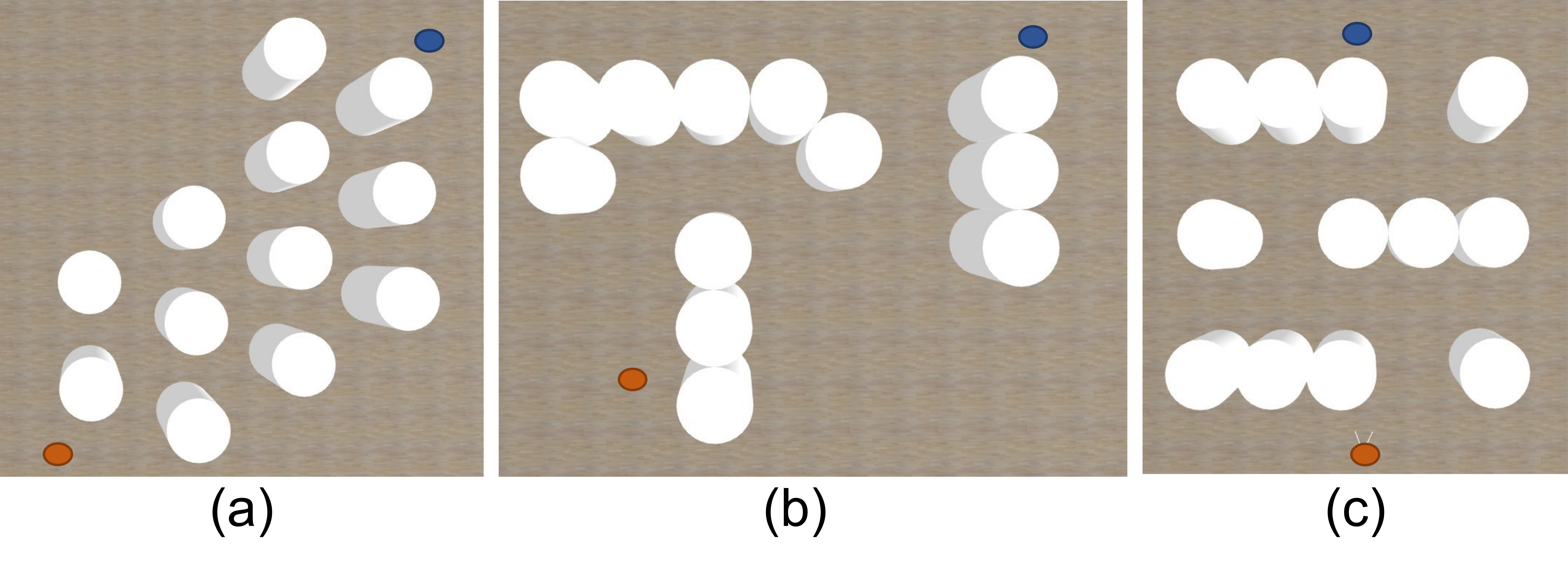}
    \caption{Experimental environments. Each environment is filled with cylinders of a radius of 0.75 meters. The quadrotor departs from the orange circle and goes to the destination depicted by the blue circle.}
    \label{fig:sim_env}
\end{figure}

In this work, we used a simulator of a quadrotor and conducted obstacle avoidance experiments.
In the simulation, the controlled quadrotor aims to reach a given goal in three types of \emph{forest-like} environments filled with cylindrical obstacles, as shown in Fig.~\ref{fig:sim_env}.
In these environments, the cylindrical obstacles are set, but the quadrotor does not know their positions in advance.
Hence, the quadrotor needs to reactively avoid them based on the sensing information from its onboard 2D LiDAR sensor.

\subsection{Formulation of Optimal Control Problem}

To apply the standard MPPI and SCP-MPPI to the navigation task, we implemented the following MPC formulation.
We first model the dynamics of the controlled quadrotor with a point mass model as follows:
\begin{align}
    {\bf x}_{t+1} = {\bf x}_{t} + {\bf u}_{t} \Delta t, \label{dynamics}
\end{align}
where ${\bf u}_t = (v_x, v_y, v_z)^ \top$ is the control input vector.
The length of the control input sequence, which is an important tuning parameter, was set to 150 in MPPI, that is 150 control points are involved in one control sequence, while it was set to 4 in SCP-MPPI.
In SCP-MPPI, these 4 control points are interpolated into a control input sequence and makes 150 control points.

To allow the quadrotor to smoothly reach at a given goal while avoiding obstacles, we design the cost function $\tilde{S}_k$ as follows:
\begin{align}
 \tilde{S}_k &= \sum_{t=0}^{T-1}\left[\Delta {\bf x}^\top  Q \Delta {\bf x} + \frac{1}{2} {\bf u}_t^ \top R {\bf u}_t   
 + \frac{w_d}{d} + C({\bf u}_t) \right], \label{cost_function}
\end{align}
where $\Delta {\bf x} = {\bf x}_t-{\bf x}_d$ represents the position error between the current position to the goal, $d$ denotes the distance to the nearest obstacle, and  $Q$, $R$, and $w_d$ are weight parameters, respectively.
$C(\cdot)$ is an indicator function of constraint conditions and is expressed as follows:
\begin{align}
    & C({\bf u}_t) = \nonumber\\ 
    & \quad\left\{
    \begin{matrix}
      1 + w_v (\|  {\bf u}_t \|_{2} - \|  {\bf u}_{\mathrm{max}} \|_{2} )& \text{if } \|  {\bf u}_t \|_{2} > \|  {\bf u}_{\mathrm{max}} \|_{2} \\
      0 & \text{otherwise } \\
    \end{matrix}
    \right.,
\end{align}
where $w_v$ are weight parameters and ${\bf u}_{\mathrm{max}}$ is a constraint condition.
We then define the optimality likelihood $p(\Delta\tilde{U})$ for the SVGD in (\ref{SVGD_}) as follows
\begin{align}
    & p(\Delta\tilde{U}) = ((\tilde{S}_k -\beta) + 1000)^{-1}, \text{ where } \beta = \min_{\Delta\tilde{U}}\tilde{S}_k,
\end{align}
and the kernel function in equation~(\ref{SVGD_}) is the following RBF kernel in this paper as
\begin{align}
    k(\Delta\tilde{U}^{(j)},~\Delta\tilde{U}) &= \exp \left(-  \frac{\|\Delta\tilde{U}^{(j)},~\Delta\tilde{U}\|_2^2}{\sigma}\right),
\end{align}
where $\sigma$ is denoted as
\begin{align}
    \sigma &= \frac{\text{median}(\| \Delta\tilde{U}^{(0)}\|^2,~\| \Delta\tilde{U}^{(1)}\|^2,~ \ldots, ~\| \Delta\tilde{U}^{(K-1)}\|^2 )} {\log K},
\end{align}
where ${\rm median}(\cdot)$ takes the median of the given value set.
Since the cost function shown in equation~(\ref{cost_function}) is not differentiable, we used a numerical differentiation for the gradient shown in equation~(\ref{SVGD_}).

\subsection{Simulation Results}

\begin{table}[t]
    \centering
    \caption{Simulation results for each environment\\ (SR: Success Rate, FT: Flight Time, AS: Average Speed)}
    \begin{minipage}{1.0\linewidth}
        \centering
        % \subcaption{Result in environment~(a)}
        \begin{tabular}{c|c|c|c}
        \toprule
        Environment (a) & SR [$\%$] & FT [s] & AS [m/s] \\
        \midrule
        SCP-MPPI & 100 & 28.3 & 0.686 \\
        MPPI ($K$ = 50) & 100 & 31.7 & 0.541 \\
        MPPI ($K$ = 1000) & 100 & 27.3 & 0.606 \\
        %SI-MPPI
        SCP-MPPI w/o SVGD ($K$ = 50) & 100 & 28.7 & 0.668 \\
        %SI-MPPI 
        SCP-MPPI w/o SVGD ($K$ = 1000) & 100 & 28.1 & 0.694 \\
        \bottomrule
        \end{tabular}
        \label{table:algorithm_environment(a)}
    \end{minipage}
    
    \begin{minipage}{1.0\linewidth}
        \centering
        % \subcaption{Result in environment~(b)}
        \begin{tabular}{c|c|c|c}
        \toprule
        Environment (b) & SR [$\%$] & FT [s] & AS [m/s] \\
        \midrule
        SCP-MPPI & 60 & 23.7 & 0.707 \\
        MPPI ($K$ = 50) & 0 & - & - \\
        MPPI ($K$ = 1000) & 0 & - & - \\
        %SI-MPPI
        SCP-MPPI w/o SVGD ($K$ = 50) & 20 & 23.9 & 0.685 \\
        %SI-MPPI
        SCP-MPPI w/o SVGD ($K$ = 1000) & 100 & 22.0 & 0.738 \\
        \bottomrule
        \end{tabular}
        \label{table:algorithm_environment(b)}
    \end{minipage}
    \begin{minipage}{1.0\linewidth}
        \centering
        % \subcaption{Result in environment~(c)}
        \begin{tabular}{c|c|c|c}
        \toprule
        Environment (c) & SR [$\%$] & FT [s] & AS [m/s] \\
        \midrule
        SCP-MPPI & 20 & 21.3 & 0.704 \\
        MPPI ($K$ = 50) & 0 & - & - \\
        MPPI ($K$ = 1000) & 0 & - & -  \\
        %SI-MPPI
        SCP-MPPI w/o SVGD ($K$ = 50) & 0 & - & - \\
        %SI-MPPI
        SCP-MPPI w/o SVGD ($K$ = 1000) & 70 & 19.7 & 0.735 \\
        \bottomrule
        \end{tabular}
        \label{table:algorithm_environment(c)}
    \end{minipage}
    
\end{table}

Based on these conditions, we conducted simulation experiments comparing SCP-MPPI, %SI-MPPI
SCP-MPPI w/o SVGD, and the standard MPPI.
SCP-MPPI was tested with 50 sample trajectories, while SCP-MPPI w/o SVGD and the vanilla MPPI were tested with 50 and 1000 sample trajectories. 
The results are summarized in Table~\ref{table:algorithm_environment(a)}, where $K$ represents the number of sampled trajectories.
The criteria for failure were defined as that the quadrotor collided against the obstacles or got stuck in local minima. 

In SCP-MPPI, the success rate of obstacle avoidance has improved. 
This indicates that SCP-MPPI can generate diverse sample trajectories, allowing the controller to predict trajectories for obstacle avoidance. 
From the results of flight time and average speed, %SI-MPPI
SCP-MPPI w/o SVGD predicts faster control inputs with fewer samples compared to the standard MPPI, enabling exploration of more distant points and significantly reducing the probability of falling into local minima.

However, as shown in Table~\ref{table:algorithm_environment(a)}, the flight time with 1000 samples is slower.
This could be attributed to modeling errors introduced by using the velocity dynamics model shown in equation~(\ref{dynamics}).
In environment (a), the quadrotor maneuvers around obstacles, leading to drift and an increase in flight time.
One possible solution is to use alternative dynamics models such as an acceleration dynamics model.

Furthermore, SCP-MPPI demonstrates the capability to generate trajectories without getting stuck in local minima even with fewer samples, emphasizing its ability to avoid obstacles.
This is due to SCP-MPPI's inherent property of transporting trajectories in a spread-out manner, resulting in reduced chances of collision with obstacles or falling into local minima.

Nevertheless, the superior performance of SCP-MPPI comes at the cost of increased computational overhead.
In our implementation, the SCP-MPPI controller runs at 10 Hz, while MPPI and %SI-MPPI
SCP-MPPI w/o SVGD controllers run at 34~Hz. 
All simulations were conducted on a desktop computer equipped with a  Intel(R) Core(TM) i9-10980XE CPU @ 3.00~GHz.

The main reasons for the increased computational cost of SCP-MPPI are associated with trajectory transportation in each iteration of the SVGD algorithm shown in equations~(\ref{SVGD_transport}), (\ref{SVGD_}) and numerical differentiation of the cost function.
Possible solutions to address this issue include parallelizing computations on GPUs for speedup and switching to analytical differentiation of the cost function when possible.

\section{Conclusion}
\label{sec:Conclusion}

In this paper, we proposed the sparse control points model predictive path integral (SCP-MPPI), which outperforms the conventional MPPI method in challenging reactive navigation tasks.
Our proposed approach employs flexible control input approximation using spline curves, enabling effective sampling even with sparse control points.
Furthermore, we introduced an algorithm for high-precision inference, representing the posterior distribution as a set of multiple particles and optimizing it using Stein Variational Gradient Descent (SVGD).
This approach allows us to improve sampling efficiency by transporting samples based on gradient-based information, in contrast to pure Monte Carlo sampling methods.
This enhancement increases the feasibility of interpolated samples.
The main contribution of this work lies in achieving smooth and collision-free motion even with sparse control points by combining spline interpolation and the SVGD algorithm.

\section*{ACKNOWLEDGMENT}

This work was supported by KAKENHI under Grant 23K03773.

%%%%%%%%%%%%%%%%%%%%%%%%%%%%%%%%%%%%%%%%%%%%%%%%%%%%%%%%%%%%%%%%%%%%%%%%%%%%%%%%

%\begin{thebibliography}{99}

\balance
\bibliographystyle{unsrt}
\bibliography{mppi.bib}

%\end{thebibliography}

\end{document}